\documentclass[letterpaper, 10 pt, journal, twoside]{ieeetran} 

\IEEEoverridecommandlockouts

\usepackage[noadjust]{cite}

\usepackage{graphicx}
\usepackage{amsmath,amssymb,amsfonts,amscd}
\usepackage{xparse}
\usepackage{subcaption}
\usepackage[keeplastbox]{flushend}

\makeatletter
\let\NAT@parse\undefined
\makeatother
\usepackage[colorlinks]{hyperref}
\usepackage{cleveref}

\usepackage{url}

\usepackage{booktabs}
\usepackage[para]{threeparttable}

\NewDocumentCommand\bbm{}{ \begin{bmatrix} }
\NewDocumentCommand\ebm{}{ \end{bmatrix} }

\NewDocumentCommand\Vector{m}{ \boldsymbol{\mathbf{#1}} }
\NewDocumentCommand\Matrix{m}{ \boldsymbol{\mathbf{#1}} }

\NewDocumentCommand\Norm{m}{\left\Vert#1\right\Vert }

\NewDocumentCommand\Real{}{ \mathbb{R} }

\NewDocumentCommand\LieGroupSE{m}{ \mathrm{SE} (#1) }

\NewDocumentCommand\Image{}{\Matrix{I}}
\NewDocumentCommand\InvariantImage{}{\Matrix{F}}

\NewDocumentCommand\CNNLoss{}{\mathcal{L}}

\NewDocumentCommand\MatcherNetParams{}{\Vector{\theta}}

\NewDocumentCommand\EncoderNetParams{}{\Vector{\phi}}
\NewDocumentCommand\TransformerNetParams{}{\Vector{\psi}}
\NewDocumentCommand\TransformParams{}{\Vector{\eta}}

\NewDocumentCommand\Matcher{}{M}
\NewDocumentCommand\MatcherNet{}{\mathcal{M}_{\MatcherNetParams}}

\NewDocumentCommand\EncoderNet{}{\mathcal{E}_{\EncoderNetParams}}
\NewDocumentCommand\TransformerNet{}{\mathcal{T}_{\TransformerNetParams}}

\graphicspath{{figs/}}

\newcommand{\icratitle}{Learning Matchable Image Transformations for Long-term Metric Visual Localization}
\newcommand{\shorttitle}{\icratitle}
\title{\icratitle}

\author{Lee Clement$^{1}$, Mona Gridseth$^{2}$, Justin Tomasi$^{1}$ and Jonathan Kelly$^{1}$%
\thanks{Manuscript received: September 10, 2019; Revised December 2, 2019; Accepted December 31, 2019.}
\thanks{This paper was recommended for publication by Editor Cesar Cadena Lerma upon evaluation of the Associate Editor and Reviewers' comments.} 
\thanks{ $^{1}$Space and Terrestrial Autonomous Robotic Systems (STARS) Lab, University of Toronto Institute for Aerospace Studies (UTIAS), Canada. }%
\thanks{ $^{2}$Autonomous Space Robotics Lab (ASRL), UTIAS, Canada. }
\thanks{ {\tt <firstname>.<lastname>@robotics.utias.utoronto.ca.} }%
\thanks{Digital Object Identifier (DOI): see top of this page.}
}

\begin{document} 

\maketitle 


\markboth{IEEE Robotics and Automation Letters. Preprint Version. Accepted December 2019}
{Clement \MakeLowercase{\textit{et al.}}: \shorttitle}  



\begin{abstract}
	Long-term metric self-localization is an essential capability of autonomous mobile robots, but remains challenging for vision-based systems due to appearance changes caused by lighting, weather, or seasonal variations.
	While experience-based mapping has proven to be an effective technique for bridging the `appearance gap,' the number of experiences required for reliable metric localization over days or months can be very large, and methods for reducing the necessary number of experiences are needed for this approach to scale.
	Taking inspiration from color constancy theory, we learn a nonlinear RGB-to-grayscale mapping that explicitly maximizes the number of inlier feature matches for images captured under different lighting and weather conditions, and use it as a pre-processing step in a conventional single-experience localization pipeline to improve its robustness to appearance change.
  We train this mapping by approximating the target non-differentiable localization pipeline with a deep neural network, and find that incorporating a learned low-dimensional context feature can further improve cross-appearance feature matching.
	Using synthetic and real-world datasets, we demonstrate substantial improvements in localization performance across day-night cycles, enabling continuous metric localization over a 30-hour period using a single mapping experience, and allowing experience-based localization to scale to long deployments with dramatically reduced data requirements.
\end{abstract}


\section{Introduction}
\IEEEPARstart{S}{elf-localization} is an essential capability of autonomous mobile robots, and localization algorithms based on inexpensive commercial vision sensors have become useful and widespread.
Despite this success, long-term \emph{metric} localization, where the goal is to continuously estimate the \mbox{6-dof} pose of the vehicle with respect to a visual map, remains challenging in the presence of appearance change caused by illumination variations over the course of a day, changes in weather conditions, or seasonal shifts.
This difficulty is largely due to simplifying assumptions such as brightness constancy and feature descriptor invariance that, when violated, cause visual localization systems to fail.
Ideally, we would like our systems to function across this `appearance gap', immune to variations in environmental conditions.

Long-term maps based on multiple visual `experiences' of an environment have proven to be effective tools for metric localization through daily and seasonal appearance change~\cite{Churchill2013-ng,Linegar2015-xs,Paton2016-bz,Paton2018-qa}.
In~\cite{Paton2016-bz}, consecutive visual experiences are recorded in a spatio-temporal pose graph, and localization against a privileged mapping experience proceeds by recalling a relevant experience and tracing through a chain of relative transformations in the graph.
This process is often aided by a prior on the vehicle's topological location in the graph, whether from dead reckoning, place recognition, or GNSS, which serves to limit the number of candidate vertices for metric localization.
However, the number of intermediate `bridging' experiences required for reliable long-term localization can be very large, and methods for compressing experience graphs are necessary for this approach to scale.

Recent work in~\cite{Clement2018-vm,Porav2018-xb} has explored deep image-to-image translation~\cite{Isola2017-uc,Zhu2017-qc} as a means of directly bridging the `appearance gap' and localizing with fewer experiences.
However these methods rely at least in part on well-aligned training images, which are difficult to obtain at scale in the real world.
Moreover, the losses used to train these models are not explicitly connected to a target localization pipeline, and provide few assurances that the learned image transformations will ultimately improve localization performance.

\begin{figure*}
  \centering
  \includegraphics[width=0.97\textwidth]{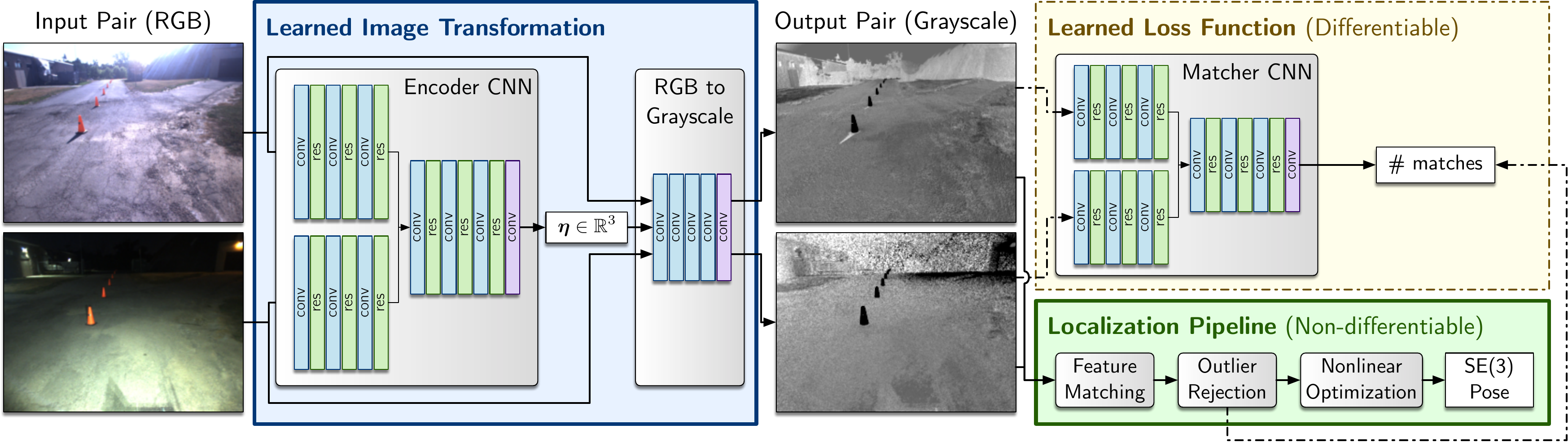}
  \caption{We learn an image transformation that improves visual feature matching performance over day-night cycles by maximizing the response of a differentiable proxy network trained to predict the number of inlier feature matches returned by a conventional non-differentiable feature detection/matching algorithm. The learned transformation can then be used as a pre-processing step in a visual localization pipeline to improve its robustness to appearance change.}
  \label{fig:pipeline}
  \vspace{-12pt}
\end{figure*}

We address these limitations by learning an image transformation optimized for a given combination of localization pipeline, sensor, and operating environment.
Rather than translating between arbitrary appearance conditions, we learn to map images to a \emph{maximally matchable} representation (i.e., one which maximizes the number of inlier feature matches) for a given feature detection/matching algorithm.
Specifically, we learn a drop-in replacement for the standard RGB-to-grayscale colorspace mapping used to pre-process RGB images for use with conventional feature detection/matching algorithms, which typically operate on single-channel images (\Cref{fig:pipeline}).
This formulation builds upon prior work on color constancy theory~\cite{Ratnasingam2010-cr}, does not require well-aligned images for training, and naturally admits a self-supervised training approach as training targets can be generated on the fly by the localization pipeline.
Our main contributions are:
\begin{enumerate}
  \item a technique for improving the robustness of a conventional visual localization pipeline to appearance change using a learned image pre-processing step;
  \item a method for optimizing the performance of a non-differentiable localization pipeline by approximating the pipeline using a deep neural network;
  \item experimental results on synthetic and real long-term vision datasets showing that our method enables continuous 6-dof metric visual localization across day-night cycles using a single mapping experience; and
  \item an open-source implementation of our method using PyTorch~\cite{paszke2017automatic}.\footnote{\url{github.com/utiasSTARS/matchable-image-transforms}}
\end{enumerate}

\section{Related Work}
Appearance robustness in metric visual localization has previously been studied from the perspective of illumination invariance, with methods such as~\cite{Clement2017-gx, Corke2013-hl, McManus2014-op, Paton2017-fi} making use of hand-engineered image transformations to improve feature matching over time for a given feature detector and descriptor.
Similarly, affine models and other simple analytical transformations have been used to improve the robustness of direct visual localization to illumination change~\cite{Engel2015-il,Park2017-zx}.
Other approaches such as~\cite{McManus2015-vj,Linegar2016-cn,Krajnik2017-tz,Zhang2018-ib} have focused on learning feature descriptors that are robust to certain types of appearance change in autonomous route following applications.
However, \cite{McManus2015-vj,Linegar2016-cn} produce correspondences that are only weakly localized, and \cite{Krajnik2017-tz,Zhang2018-ib} require sets of true and false point correspondences to train feature descriptors, which are challenging to obtain at scale over long periods.

Deep image-to-image translation~\cite{Isola2017-uc,Zhu2017-qc} has recently been applied to the problem of metric localization across appearance change.
In~\cite{Gomez-Ojeda2018-ai} the authors train a convolutional encoder-decoder network to enhance the temporal consistency of image streams captured in environments with high dynamic range.
Here the main source of appearance change is the camera itself as it automatically modulates its imaging parameters in response to the local brightness of a static environment.
Other work has tackled the problem of localization across \emph{environmental} appearance change, with \cite{Clement2018-vm} learning a many-to-one mapping onto a privileged appearance condition and \cite{Porav2018-xb} learning multiple pairwise mappings between appearance categories such as day and night.
Image-to-image translation has also been applied to the related task of appearance-invariant place recognition~\cite{Latif2018-ui,Anoosheh2019-fc}, which typically relies on patch matching or whole-image statistics to identify images corresponding to nearby physical locations rather than estimating the 6-dof pose of the vehicle.
While \cite{Porav2018-xb,Gomez-Ojeda2018-ai} include loss terms to maximize gradient information, these heuristics are not directly tied to the performance of the localization pipeline.
Moreover, \cite{Clement2018-vm,Porav2018-xb,Gomez-Ojeda2018-ai} require well-aligned training images exhibiting appearance variation, which are difficult to obtain at scale in the real world, and it is not clear how categorical appearance mappings such as \cite{Porav2018-xb,Latif2018-ui,Anoosheh2019-fc} should be applied to continuous appearance change in long-term deployments.

Surrogate-based methods for approximating computationally expensive or non-differentiable objective functions are common in the numerical optimization literature~\cite{Koziel2011}.
Neural network surrogates in particular have found applications in a variety of domains including computer graphics~\cite{Grzeszczuk:1998:FNN:3009055.3009178} and computational oceanography~\cite{VANDERMERWE2007462}, where high-fidelity physics simulations are available but expensive to compute.
Our method of learning a differentiable loss function is similar in spirit to Generative Adversarial Networks (GANs)~\cite{Goodfellow2014-df} in that a complex discriminator/loss function is trained using a comparatively simple analytical loss function.
It also bears resemblances to perceptual losses~\cite{Johnson2016-lu}, where the loss function is derived from the feature activations of a network trained on a proxy task such as image classification.

\section{Learning Matchable Colorspace Transformations}
Our goal in this work is to learn a nonlinear transformation $f: \Real^3 \rightarrow \Real $ mapping the RGB colorspace onto a grayscale colorspace that explicitly maximizes a chosen performance metric of a vision-based localization pipeline.
We investigate two approaches to formulating such a mapping: 1) a single function to be applied as a pre-processing step to all incoming images, similarly to \cite{Clement2017-gx,McManus2014-op,Paton2017-fi}; and 2) a parametrized function tailored to the specific image pair to be used for localization, where the parameters of this function are derived from the images themselves.
Additionally, the functional form of either mapping may be specified analytically (e.g., from physics) or learned from data using a function approximator such as a neural network.

In order to find an optimal colorspace transformation for a given application, we require an appropriate objective function to optimize, which should ideally be tied to the performance of the target localization pipeline.
An intuitive choice of objective could be to directly minimize pose estimation error for the entire pipeline relative to ground truth if it is available.
In the absence of accurate ground truth data, we might instead choose to maximize the number or quality of feature matches in the front-end of a feature-based localization pipeline.
We adopt the latter approach in this work, since high-quality 6-dof ground truth is difficult to obtain over long time scales.

Although it is straightforward to choose a target performance metric to optimize, the most commonly used localization front-ends in robotics rely on non-differentiable components such as stereo matching, nearest-neighbors search, and RANSAC~\cite{Fischler1981-ue}, which are incompatible with the gradient-based optimization schemes commonly used in deep learning.
In this work we \emph{learn} an objective function by training a deep convolutional neural network (CNN) to act as a \emph{differentiable proxy} to the localization front-end.
Specifically, we train a siamese CNN to predict the number of inlier feature matches for a given image pair, where the training targets are generated using a conventional non-differentiable feature detector/matcher algorithm based on \texttt{libviso2} features~\cite{Geiger2011-xe}.
This proxy network can then be used to define a fully differentiable objective function, allowing us to train a nonlinear colorspace mapping using gradient-based methods.
Finally, the trained image transformation can be used as a pre-processing step in a conventional visual localization pipeline, enabling it to operate more reliably under appearance change.
\Cref{fig:pipeline} summarizes our full data pipeline pictorially.

\subsection{Differentiable Matcher Proxy} 
\label{sec:matcher}

We consider the task of training a CNN $\MatcherNet$, with parameters $\MatcherNetParams$, to predict the number of inlier feature matches returned by a non-differentiable feature detector/matcher $\Matcher$ for a given image pair $ (\Image_1, \Image_2) $.
This training objective is a convenient choice for our intended application as it is closely tied to the ability of our visual localization pipeline to operate across appearance change, however it is not by any means the only choice.
For example, we could also train a CNN to predict a measure of localization accuracy such as geodesic distance from ground truth on the $\LieGroupSE{3}$ manifold, similarly to the estimator correction framework proposed in~\cite{Peretroukhin2018-qc}.
Importantly, this formulation admits a self-supervised training approach as training targets can be generated automatically by $\Matcher$.

\Cref{fig:pipeline} (right-hand side) summarizes the training setup for this task.
A pair of single-channel images is fed into a conventional feature detection/matching algorithm (e.g.,  SURF~\cite{Bay2008-zi}, ORB~\cite{Rublee2011-hd}, or \texttt{libviso2}~\cite{Geiger2011-xe}), and a summary statistic is computed such as the quantity of RANSAC-filtered inlier feature matches.
This summary statistic forms the training target for a CNN whose task is to predict the statistic for the same image pair.
Given enough training pairs, the network should learn a set of convolutional filters that correspond to the types of features and patterns that best predict the performance of $\Matcher$ in a given environment.
Critically, the proxy network is fully differentiable and can provide a gradient signal to train a nonlinear image transformation.

\begin{figure}
  \centering
  \includegraphics[width=0.9\columnwidth]{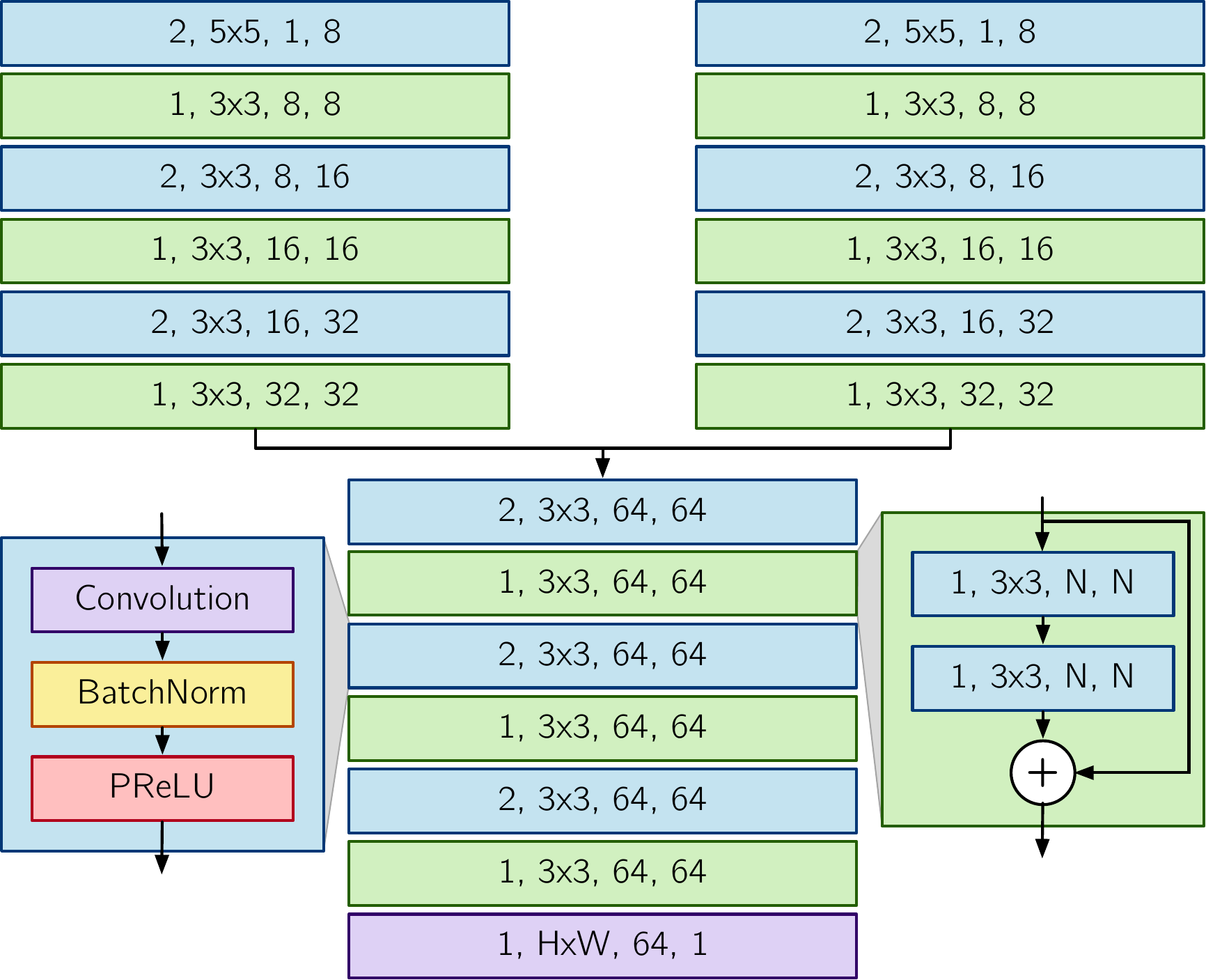}
  \caption{Network architecture for $\MatcherNet$. Each block is denoted by stride, kernel size, input channels, and output channels. The left and right branches share weights. The final output is produced by a fully-connected layer (implemented as a convolution operator), which projects the feature map to a scalar value.}
  \label{fig:matcher_net}
  \vspace{-12pt}
\end{figure}

Our matcher proxy network (\Cref{fig:matcher_net}) is a siamese network built from convolutional and residual~\cite{He2016-zg} blocks using batch normalization~\cite{Ioffe2015-bs} and PReLU non-linearities~\cite{He2015-zg}.
Each image in the input pair is processed by one of two feature detection branches, which share weights to ensure that both images are mapped onto a common feature space.
The outputs of the feature detection branches are concatenated along the channel dimension to be further processed by the remainder of the network.
Each non-residual convolution block downsamples the feature map by a factor of two, allowing for salient features to be learned at multiple scales.
The final output is produced by a fully-connected layer, which projects the feature map to a scalar value.
We train $\MatcherNet$ to fit $\Matcher$ in a least-squares sense by minimizing the mean squared error of the predicted match counts for a minibatch of $N$ image pairs:
\begin{align} \label{eq:matcher_loss}
  \CNNLoss(\MatcherNetParams) & = \frac{1}{N} \sum_{i=1}^N \left( \MatcherNet(\Image_1^i, \Image_2^i) - \Matcher(\Image_1^i, \Image_2^i) \right)^2.
\end{align}

\subsection{Physically Motivated Transformations}
\label{sec:logrgb}
Prior work in \cite{Ratnasingam2010-cr} has shown that under the assumptions of a single black-body illuminant and an infinitely narrow sensor response function, an appropriately weighted linear combination of the log-responses of a three-channel (e.g., RGB) camera represents a projection onto an invariant one-dimensional chromaticity space that is independent of both the intensity and color temperature of the illuminant, and depends only on the imaging sensor and the materials in the scene:
\begin{align} \label{eq:color_constant}
  \InvariantImage^i_j = \log \Image^i_j({\lambda_2}) - \alpha \log\Image^i_j({\lambda_1}) - \beta \log\Image^i_j({\lambda_3}),
\end{align}
where $ \Image^i_j({\lambda_k}) $ is the image of sensor responses at wavelength $\lambda_k$, the weights $(\alpha, \beta)$ are subject to the constraints
\begin{align} \label{eq:color_constant_constraints}
  \frac{1}{\lambda_2} & = \frac{\alpha}{\lambda_1} + \frac{\beta}{\lambda_3} & \mathrm{and} &  & \beta & = (1-\alpha),
\end{align}
and the indices $k$ are chosen such that $\lambda_1 < \lambda_2 < \lambda_3$ (i.e., red, green and blue channels, respectively).

The image formed from this pixel-wise linear combination of log-responses can then be rescaled to produce a valid grayscale image that can be further processed by a localization pipeline.
Grayscale images generated using this procedure are somewhat resistant to variations in lighting and shadow, and have been shown to improve stereo localization quality in the presence of shadows and changing daytime lighting conditions \cite{Clement2017-gx,McManus2014-op,Paton2017-fi}, but have not been successful in adapting to nighttime navigation with headlights.

Given the constraints defined by~\Cref{eq:color_constant_constraints}, the weights $(\alpha, \beta)$ are completely specified as a function of the imaging sensor.
However, in practice, these constraints are relaxed and the parameters $(\alpha, \beta)$ are tuned to a specific environment, sensor, and feature matcher, where the theoretical values do not perform optimally.
Indeed, \cite{Clement2017-gx,Paton2017-fi} used two sets of parameters tuned to maximize the stability of SURF features~\cite{Bay2008-zi} in regions where grassy or sandy materials dominate.

We argue that environmental appearance is best thought of as continuous rather than categorical, and that a better approach to selecting the transformation parameters should take into account the content of the specific scene being imaged, rather than using the same parameters at every location within a large and potentially heterogeneous operating environment.
Accordingly, we train a second encoder network $\EncoderNet$, with parameters $\EncoderNetParams$, to predict the optimal values of the transformation parameters (i.e., which yield the most inlier feature matches) for a given RGB image pair.

Furthermore, we relax the constraints in~\Cref{eq:color_constant_constraints} and generalize~\Cref{eq:color_constant} to be of the form
\begin{align} \label{eq:color_constant_generalized}
  \InvariantImage^i_j & = \alpha^i \log \Image^i_j({\lambda_1}) + \beta^i \log\Image^i_j({\lambda_2}) + \gamma^i \log\Image^i_j({\lambda_3}),
\end{align}
where the parameters are computed for the $i^\mathrm{th}$ image pair as
\begin{align} \label{eq:encoder_output}
  \TransformParams^i & = \bbm \alpha^i & \beta^i & \gamma^i \ebm^T  = \EncoderNet(\Image^i_1, \Image^i_2),
\end{align}
and~\Cref{eq:color_constant_generalized} is applied to both $\Image^i_1$ and $\Image^i_2$ using the same set of parameters.
Due to the need to rescale $\InvariantImage^i_j$ to form a valid single-channel image, a degree of freedom in $\TransformParams^i$ is lost and $(\alpha, \beta, \gamma)$ represent the relative mixing proportions of the three color channels.
We enforce $\Norm{ \TransformParams^i }_1 = 1$ using a normalization layer to ensure a consistent range of outputs.

Our encoder network $\EncoderNet$ follows a similar siamese architecture to $\MatcherNet$, but takes pairs of 3-channel images as inputs and outputs a 3-dimensional vector.
We train $\EncoderNet$ to maximize the mean number of inlier feature matches as predicted by $\MatcherNet$, or equivalently, to minimize its negation:
\begin{align} \label{eq:encoder_loss}
  \CNNLoss(\EncoderNetParams) & = -\frac{1}{N} \sum_{i=1}^N \MatcherNet(\InvariantImage_1^i, \InvariantImage_2^i),
\end{align}
where $\InvariantImage_1^i, \InvariantImage_2^i$ are computed from input RGB images $\Image_1^i, \Image_2^i$ using \Cref{eq:color_constant_generalized,eq:encoder_output}.

Rather than rescaling using the minimum and maximum response of each output image, we rescale by the joint mean $ \mu^i $ and standard deviation $ \sigma^i $ of the output pair and apply a clamping operation to map the output onto the range $ [0, 1] $:
\begin{align} \label{eq:rescale_instancenorm}
  \InvariantImage^i_j & \leftarrow \frac{1}{2} \left[ \frac{\InvariantImage^i_j - \mu^i }{3 \sigma^i} \right]_{-1,1} + \frac{1}{2},
\end{align}
where we have used the notation $\left[ \cdot \right]_{a,b} = \min(\max(\cdot, a), b)$.
This rescaling scheme allows the model to saturate parts of the output images while still using the full range of valid pixel values.
Moreover, it avoids introducing significant sparsity in the gradients through the $\min(\cdot)$ and $\max(\cdot)$ operators, which improves the flow of gradient information during training.


\subsection{Learned Nonlinear Transformations}
\label{sec:learned}
While the assumption of a single black-body illuminant in~\cite{Ratnasingam2010-cr} is reasonable for daytime navigation where the dominant light source is the sun, it does not hold in many common navigation scenarios such as nighttime driving with headlights.
Moreover, the assumption of an infinitely narrow sensor response is unrealistic for real cameras.
As an alternative to the physically motivated colorspace transformation outlined in \Cref{sec:logrgb}, we investigate the possibility of learning a bespoke nonlinear mapping that maximizes matchability for a particular combination of imaging sensor, estimator and environment.
We parametrize this mapping using a small neural network $\TransformerNet$, with parameters $\TransformerNetParams$, operating independently on each pixel of each input RGB image.
We structure $\TransformerNet$ as a multilayer perceptron (MLP) implemented using $1 \times 1$ convolutions and PReLU nonlinearities.

We consider two versions of this MLP-based transformation, both with and without incorporating an additional pairwise context feature obtained from encoder network $\EncoderNet$ using \Cref{eq:encoder_output}.
In the case where $\EncoderNet$ is used, the input to $\TransformerNet$ becomes the concatenation of the input RGB image and the parameters $\TransformParams^i$ along the channel dimension, and the first convolutional layer of $\TransformerNet$ is modified accordingly.
We train $\TransformerNet$ and $\EncoderNet$ (if used) jointly by minimizing a similar loss function to \Cref{eq:encoder_loss}, where in place of \Cref{eq:color_constant_generalized}, we have
\begin{align} \label{eq:learned_invariant_transform}
  \InvariantImage_j^i & = \begin{cases} \TransformerNet(\Image_j^i, \TransformParams^i) & \text{if $\EncoderNet$ is used,} \\ \TransformerNet(\Image_j^i) & \text{if $\EncoderNet$ is not used}.  \end{cases}
\end{align}
Similarly to the physically motivated transformations described in~\Cref{sec:logrgb}, we rescale $\InvariantImage_j^i$ to fill the range of valid grayscale values by applying~\Cref{eq:rescale_instancenorm}.

\begin{figure*}
  \centering
  \begin{subfigure}{0.32\textwidth}
    \includegraphics[width=\textwidth]{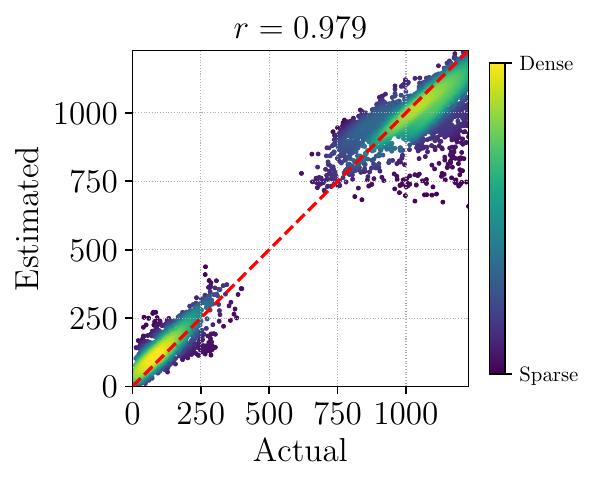}
    \caption{\texttt{VKITTI} (Morning vs. Sunset)}
    \label{fig:matcher_target_est:vkitti_morning-sunset}
  \end{subfigure}
  ~
  \begin{subfigure}{0.31\textwidth}
    \includegraphics[width=\textwidth]{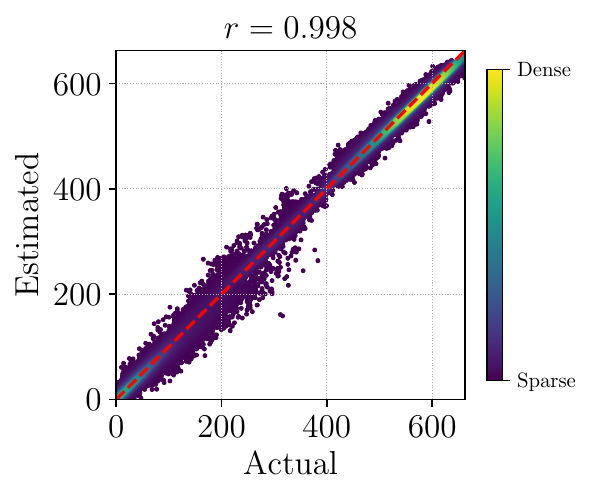}
    \caption{\texttt{InTheDark}}
    \label{fig:matcher_target_est:inthedark}
  \end{subfigure}
  ~
  \begin{subfigure}{0.31\textwidth}
    \includegraphics[width=\textwidth]{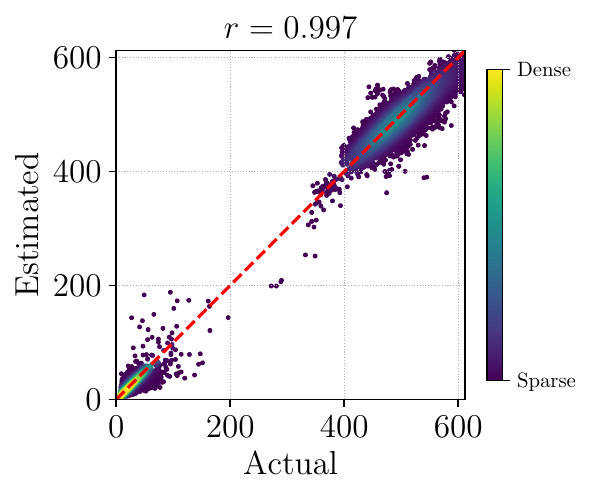}
    \caption{\texttt{RobotCar}}
    \label{fig:matcher_target_est:oxford}
  \end{subfigure}
  \caption{Estimated vs. actual match counts for $\MatcherNet$ after ten epochs of pre-training on each dataset, colour-coded by relative density. Match counts are aggregated over all test sequences and include self-matches. The dashed red line corresponds to perfect agreement of $\MatcherNet$ with \texttt{libviso2}. In each case, the match count predictions produced by $\MatcherNet$ are very strongly correlated with the true match counts.}
  \label{fig:matcher_target_est}
\end{figure*}

\begin{figure*}
  \centering
  \begin{subfigure}{0.95\textwidth}
    \includegraphics[width=\textwidth]{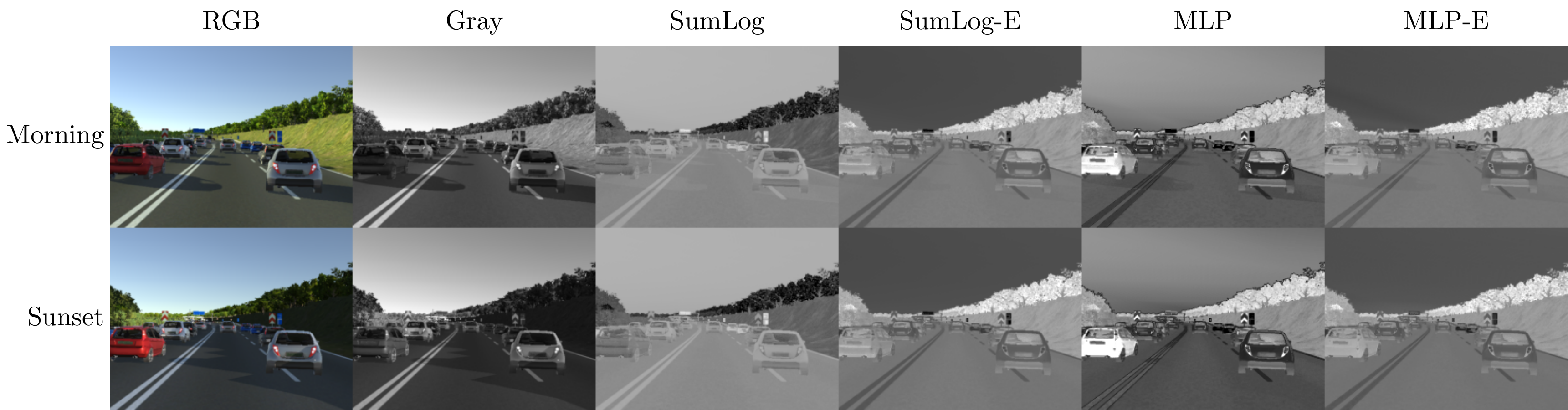}
    \caption{\texttt{VKITTI/0020} (Sunset to Morning, cropped)}
    \label{fig:vkitti_0020_grid}
  \end{subfigure}
  ~
  \begin{subfigure}{0.95\textwidth}
    \includegraphics[width=\textwidth]{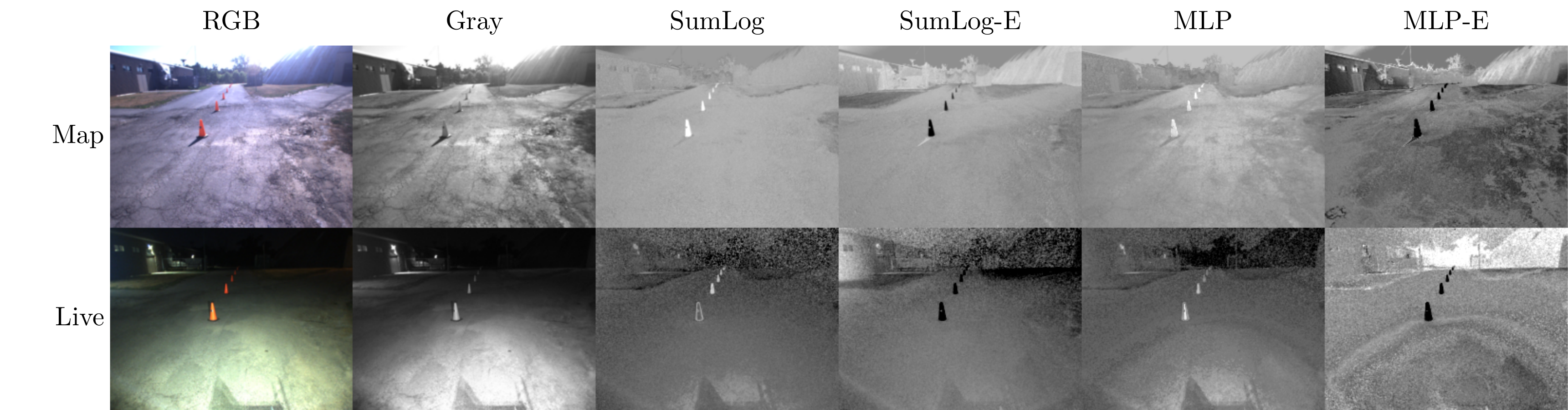}
    \caption{\texttt{InTheDark/0041} (Night to Day)}
    \label{fig:inthedark_000041_grid}
  \end{subfigure}
  \caption{Sample input RGB pairs and corresponding outputs of each RGB-to-grayscale transformation.}
  \label{fig:output_grid}
  \vspace{-12pt}
\end{figure*}

\section{Experiments}
We conducted experiments on synthetic and real-world long-term vision datasets to validate and compare each approach.
Specifically, we evaluated the ability of our matcher proxy network to capture the performance of \texttt{libviso2} feature matching across viewpoint and appearance changes, as well as the effect of each image transformation on feature matching and localization performance.
When evaluating feature matching, we assumed that we had a prior on the vehicle's topological location in the map, such that we could reliably identify the nearest vertex in the pose graph.
This is typical for autonomous visual route-following systems such as~\cite{Paton2018-qa}, where the topological prior is derived by dead reckoning from a previous successful localization, or using place recognition or GNSS in the event that the system becomes lost.

We refer to the generalized color-constancy model of~\cite{Ratnasingam2010-cr} (\Cref{sec:logrgb}) as ``SumLog'' and ``SumLog-E'' , where the latter uses \Cref{eq:encoder_output} to derive the parameters $\TransformParams^i$ per image pair, and the former uses a constant $\TransformParams$ that maximizes inlier feature matches over the training set (similarly to~\cite{Clement2017-gx,Paton2017-fi}).
Analogously, we refer to the learned multilayer perceptron models (\Cref{sec:learned}) as ``MLP'' and ``MLP-E'', where the latter incorporates $\EncoderNet$ and the former does not .
We refer to the standard RGB-to-grayscale transformation as ``Gray''.\footnote{We refer specifically to the ITU-R 601-2 luma transform implemented by the \texttt{Pillow} library: $L = 0.299 R + 0.587 G + 0.114 B$.}

Training proceeds in two stages.
First, we pre-train $\MatcherNet$ using standard grayscale images.
Training labels are generated using the open-source \texttt{libviso2} library~\cite{Geiger2011-xe} to detect and match features, and the eight-point RANSAC algorithm to reject outlier matches.
Second, we train $\EncoderNet$ and/or $\TransformerNet$ using the matchability loss defined in \Cref{eq:encoder_loss}.
To ensure that $\MatcherNet$ accurately predicts feature match counts for the output images, which differ significantly from standard grayscale images, we continue to train $\MatcherNet$ in an alternating fashion using the output images at each iteration.
All models are implemented in \mbox{PyTorch}~\cite{paszke2017automatic} and trained for 10 epochs with a batch size of 8, using the Adam optimizer~\cite{Kingma2015-wl} with default parameters and a learning rate of $10^{-4}$. 
We rescale all images to a height of 192 pixels for both training and testing.

\subsection{Datasets}
We evaluated our approach using both synthetic and real-world datasets exhibiting severe illumination change.

\paragraph{Virtual KITTI} \label{sec:vkitti_dataset}
The Virtual KITTI  (\texttt{VKITTI}) dataset~\cite{Gaidon2016-by} is a synthetic reconstruction of a portion of the KITTI vision benchmark~\cite{Geiger2013-ky}, consisting of five sets of non-overlapping trajectories with \mbox{RGB-D} imagery rendered under a variety of simulated illumination conditions.
This dataset is a convenient validation tool as it provides perfect data association and a range of daytime illumination conditions.
For each trajectory, we train models using image pairs from the others.
Since each trajectory is non-overlapping, this spatial split allows us to evaluate how well our method generalizes to unseen environments.
Further, since \texttt{VKITTI} provides corresponding images from identical viewpoints, we augmented the training data to ensure generalizability to viewpoint changes and dynamic objects by associating each training image in one  condition with a window of nearby images in the other.

\paragraph{UTIAS In The Dark}
The UTIAS In The Dark (\texttt{InTheDark}) dataset~\cite{Paton2016-bz} provides stereo imagery of a 250 m outdoor loop traversed repeatedly over a 30-hour period using on-board headlights to illuminate the scene at night.
We use the multi-experience localization system in \cite{Paton2016-bz} to obtain corresponding image pairs with overlapping fields of view but non-identical poses.
We train our models using left-camera images from 66 traversals and test on 7 held-out traversals spanning a full day-night cycle (listed in \Cref{tab:match_stats}).
This temporal split allows us to evaluate how well our method generalizes to multiple unseen illumination conditions.

\paragraph{Oxford RobotCar}
We further evaluate our method using three sequences from the Oxford RobotCar (\texttt{RobotCar}) dataset~\cite{Maddern2016-ng}, captured along the same 10~km route in overcast, nighttime, and sunny conditions.\footnote{``Overcast'', ``Night'', and ``Sunny'' refer to \texttt{2014-12-09-13-21-02}, \texttt{2014-12-10-18-10-50}, and \texttt{2014-12-16-09-14-09}, respectively.}
We find corresponding images using the GNSS/INS poses, and split the trajectory into two non-overlapping segments: the first 70\% of the trajectory is used for training and the remaining 30\% is used for testing.

\begin{figure}
  \centering
  \begin{subfigure}{0.45\textwidth}
    \includegraphics[width=\textwidth]{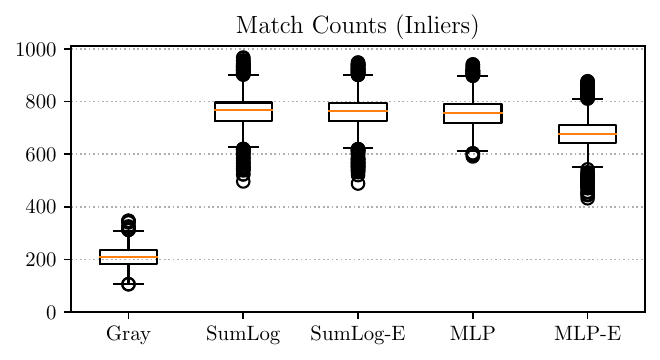}
    \caption{\texttt{VKITTI/0020} (Sunset to Morning)}
    \label{fig:vkitti_0020_matches}
  \end{subfigure}
  ~
  \begin{subfigure}{0.45\textwidth}
    \includegraphics[width=\textwidth]{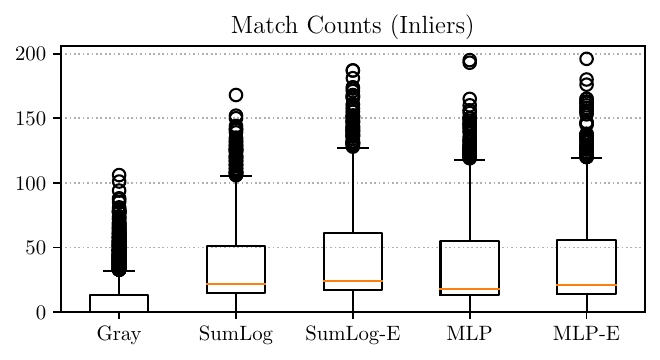}
    \caption{\texttt{InTheDark/0041} (Night to Day)}
    \label{fig:inthedark_000041_matches}
  \end{subfigure}
  \caption{Box-and-whiskers plots of inlier \texttt{libviso2} feature matches for corresponding image pairs with each RGB-to-grayscale transformation applied. Orange lines indicate the median values.}
  \label{fig:matches_box_whiskers}
  \vspace{-12pt}
\end{figure}

\begin{table}[]
  \setlength{\tabcolsep}{2.7pt}
  \centering
  \caption{Actual inlier feature matches using \texttt{libviso2} and each RGB-to-grayscale transformation. The highest mean number of matches for each sequence is highlighted in bold.}

  \begin{tabular}{@{}llccccc@{}}
    \toprule
    \multicolumn{2}{l}{}                       & \multicolumn{5}{c}{\textbf{Inlier Feature Matches $\mu (\sigma)$}}                                  \\ \cmidrule{3-7}
    \multicolumn{2}{l}{\textbf{Test Sequence}} & Gray              & SumLog           & SumLog-E           & MLP               & MLP-E              \\ \midrule
    \multicolumn{2}{l}{\texttt{VKITTI/0001}}   &                   &                    &                   &                   &                   \\
          & Sunset-Morning              & 262 (82)          & \textbf{726 (136)} & 689 (157)         & 661 (108)         & 623 (116)         \\
          & Overcast-Clone              & 444 (58)          & \textbf{790 (129)} & 767 (125)         & 747 (106)         & 770 (107)         \\ \addlinespace
\multicolumn{2}{l}{\texttt{VKITTI/0002}}   &                   &                    &                   &                   &                   \\
          & Sunset-Morning              & 240 (22)          & \textbf{812 (67)}  & 803 (70)          & 774 (65)          & 702 (62)          \\
          & Overcast-Clone              & 290 (41)          & 739 (67)           & 757 (76)          & 755 (65)          & \textbf{764 (84)} \\ \addlinespace
\multicolumn{2}{l}{\texttt{VKITTI/0006}}   &                   &                    &                   &                   &                   \\
          & Sunset-Morning              & 125 (33)          & 669 (44)           & \textbf{735 (43)} & 711 (37)          & 642 (41)          \\
          & Overcast-Clone              & 142 (33)          & 647 (39)           & \textbf{666 (43)} & 566 (47)          & 546 (38)          \\ \addlinespace
\multicolumn{2}{l}{\texttt{VKITTI/0018}}   &                   &                    &                   &                   &                   \\
          & Sunset-Morning              & 234 (53)          & \textbf{817 (36)}  & 816 (37)          & 745 (37)          & 731 (40)          \\
          & Overcast-Clone              & 311 (46)          & 548 (52)           & \textbf{555 (50)} & 450 (42)          & 486 (39)          \\ \addlinespace
\multicolumn{2}{l}{\texttt{VKITTI/0020}}   &                   &                    &                   &                   &                   \\
          & Sunset-Morning              & 210 (39)          & \textbf{762 (69)}  & 758 (71)          & 756 (62)          & 675 (69)          \\
          & Overcast-Clone              & 287 (78)          & 716 (71)           & 716 (72)          & \textbf{718 (62)} & 708 (64)          \\ \addlinespace
\multicolumn{2}{l}{\texttt{InTheDark}}     &                   &                    &                   &                   &                   \\
          & Map~~(08:54)                   & -                 & -                  & -                 & -                 & -                 \\
          & \texttt{0006} (09:46)          & \textbf{178 (48)} & 125 (56)           & 165 (51)          & 138 (52)          & 158 (53)          \\
          & \texttt{0027} (18:36)          & 9 (9)             & 52 (29)            & \textbf{57 (26)}  & 44 (25)           & 48 (28)           \\
          & \texttt{0041} (21:48)          & 10 (16)           & 33 (25)            & \textbf{39 (31)}  & 33 (29)           & 35 (29)           \\
          & \texttt{0058} (05:48)          & 99 (34)           & 95 (47)            & \textbf{114 (43)} & 97 (44)           & 113 (40)          \\
          & \texttt{0071} (09:18)          & \textbf{181 (44)} & 126 (52)           & 167 (47)          & 141 (48)          & 161 (48)          \\
          & \texttt{0083} (14:01)          & 53 (21)           & 76 (39)            & \textbf{83 (37)}  & 82 (37)           & \textbf{83 (37)}  \\
          & \texttt{0089} (16:43)          & 45 (22)           & 67 (35)            & \textbf{70 (36)}  & 63 (31)           & 68 (33)           \\ \addlinespace
\multicolumn{2}{l}{\texttt{RobotCar}}     &                   &                    &                   &                   &                   \\ 
& Overcast-Night & 11(3) & \textbf{12 (2)} & \textbf{12 (3)} & 11 (2) & 11 (2) \\
& Overcast-Sunny & 26 (12) & 25 (12) & 25 (12) & 24 (9) & \textbf{71 (41)} \\ \bottomrule
    \end{tabular}
  \label{tab:match_stats}
  \vspace{-12pt}
\end{table}

\subsection{Feature Matcher Approximation}
We train $\MatcherNet$ in a self-supervised manner to predict the number of inlier feature matches for overlapping image pairs captured under different illumination conditions from nearby but different poses.
Training labels are generated on the fly for each image pair using the open-source \texttt{libviso2} library~\cite{Geiger2011-xe} in monocular flow matching mode with default parameters, using the eight-point RANSAC algorithm to reject outlier matches.
In practice we train $\MatcherNet$ to minimize \Cref{eq:matcher_loss} over all combinations of input images.

\Cref{fig:matcher_target_est} plots actual and estimated match counts for each dataset after ten epochs of pre-training on Gray images, aggregated over all test sequences.
These include self-matches (same viewpoint, same appearance) and non-self-matches (different viewpoint, different appearance), which appear as clusters.
In each case the test-time match counts predicted by $\MatcherNet$ are strongly correlated with the true performance of \texttt{libviso2}.
This indicates that our approach generalizes well and that $\MatcherNet$ is capturing salient properties of feature matching rather than memorizing training examples.

\begin{table*}[]
  \centering
  \caption{Maximum distances travelled on dead reckoning for each test sequence of the UTIAS In The Dark dataset, based on various thresholds of inlier feature matches against the ``Map'' sequence. The best results are highlighted in bold.}
  \begin{threeparttable}  
    \begin{tabular}{@{}ll*{17}c@{}}
      \toprule
      &                                & \multicolumn{17}{c}{\textbf{Maximum Distance on Dead Reckoning (m)}}                                                                                                                                                                         \\ \cmidrule{3-19}
      &                                & \multicolumn{5}{c}{\textbf{$\ge$ 10 Inliers}}                            &  & \multicolumn{5}{c}{\textbf{$\ge$ 20 Inliers}}                            &  & \multicolumn{5}{c}{\textbf{$\ge$ 30 Inliers}}                            \\ 
  \multicolumn{2}{l}{} & G\tnote{1}            & S\tnote{2}           & S-E\tnote{2}         & M\tnote{3}          & M-E\tnote{3}        &  & G\tnote{1}            & S\tnote{2}           & S-E\tnote{2}         & M\tnote{3}          & M-E\tnote{3}        &  & G\tnote{1}            & S\tnote{2}           & S-E\tnote{2}         & M\tnote{3}          & M-E\tnote{3}        \\ \cmidrule{3-7} \cmidrule{9-13} \cmidrule{15-19}
  \multicolumn{2}{l}{\texttt{InTheDark}} &              &              &              &              &              &  &              &              &              &              &              &  &              &              &               &              &              \\
        & Map~~(08:54)                 & -            & -            & -            & -            & -            &  & -            & -            & -            & -            & -            &  & -            & -            & -             & -            & -            \\
        & \texttt{0006} (09:46)        & \textbf{0.0} & \textbf{0.0} & \textbf{0.0} & \textbf{0.0} & \textbf{0.0} &  & \textbf{0.0} & \textbf{0.0} & \textbf{0.0} & \textbf{0.0} & \textbf{0.0} &  & \textbf{0.0} & 0.1          & \textbf{0.0}  & \textbf{0.0} & \textbf{0.0} \\
        & \texttt{0027} (18:36)        & 7.5          & 0.7          & \textbf{0.0} & 0.3          & 0.3          &  & 25.9         & 3.6          & \textbf{0.7} & 2.5          & 2.2          &  & 101.1        & 10.5         & \textbf{0.7}  & 8.3          & 7.8          \\
        & \texttt{0041} (21:48)        & 14.9         & 0.5          & \textbf{0.2} & 0.9          & 1.4          &  & 46.3         & 8.4          & \textbf{3.2} & 7.2          & 7.4          &  & 104.5        & 15.1         & \textbf{10.0} & 18.6         & 15.7         \\
        & \texttt{0058} (05:48)        & \textbf{0.0} & \textbf{0.0} & \textbf{0.0} & \textbf{0.0} & \textbf{0.0} &  & \textbf{0.0} & 0.2          & \textbf{0.0} & 0.2          & \textbf{0.0} &  & \textbf{0.0} & 0.6          & 0.3           & 0.6          & \textbf{0.0} \\
        & \texttt{0071} (09:18)        & \textbf{0.0} & \textbf{0.0} & \textbf{0.0} & \textbf{0.0} & \textbf{0.0} &  & \textbf{0.0} & \textbf{0.0} & \textbf{0.0} & \textbf{0.0} & \textbf{0.0} &  & \textbf{0.0} & \textbf{0.0} & \textbf{0.0}  & \textbf{0.0} & \textbf{0.0} \\
        & \texttt{0083} (14:01)        & \textbf{0.0} & 0.2          & \textbf{0.0} & \textbf{0.0} & \textbf{0.0} &  & 0.4          & 0.5          & 0.3          & 0.2          & \textbf{0.0} &  & 2.2          & 4.0          & \textbf{0.3}  & 0.4          & 0.5          \\
        & \texttt{0089} (16:43)        & 0.3          & 0.3          & 0.2          & \textbf{0.0} & \textbf{0.0} &  & 4.8          & 1.0          & 3.7          & 1.2          & \textbf{0.8} &  & 6.1          & 6.1          & 4.7           & \textbf{2.9} & \textbf{2.9}  \\ \addlinespace
        \multicolumn{2}{l}{\texttt{RobotCar}} &              &              &              &              &              &  &              &              &              &              &              &  &              &              &               &              &              \\ 
& Map (Overcast) & - & - & - & - & - &  & - & - & - & - & - &  & - & - & - & - & - \\
& Night & 27.0 & 7.6 & \textbf{3.7} & 27.3 & 27.3 &  & 307.1 & 307.1 & \textbf{245.9} & 400.8 & 398.9 &  & 740.5 & \textbf{541.5} & 740.5 & 697.3 & 740.5 \\
& Sunny & 1.3 & 1.3 & 1.5 & 1.8 & \textbf{0.9} &  & 36.1 & 65.1 & 48.0 & 39.4 & \textbf{6.3}  &  & 129.1 & 139.8 & 109.2 & 124.3 & \textbf{16.5} \\ \bottomrule
    \end{tabular}
    \begin{tablenotes}
      \item[1] G: Gray
      \item[2] S: SumLog 
      \item[3] M: MLP 
    \end{tablenotes}
    \label{tab:inthedark:loc_succ}
  \end{threeparttable}
  \vspace{-12pt}
\end{table*}

\subsection{Feature Matching Across Appearance Change}
\Cref{fig:output_grid} shows the outputs of each image transformation for sample RGB image pairs in the \texttt{VKITTI/0020} Morning and Sunset sequences (\Cref{fig:vkitti_0020_grid}) and the challenging sequence \texttt{InTheDark/0041} (\Cref{fig:inthedark_000041_grid}).
We see that each model produced image pairs that are visually more consistent than standard Gray images, and that local illumination variations such as shadows, uneven lighting, and specular reflections were minimized by optimizing \Cref{eq:encoder_loss}.

\Cref{fig:matches_box_whiskers} visually compares the distributions of actual \texttt{libviso2} feature matches for each transformation, while \Cref{tab:match_stats} summarizes the results numerically.
Each model significantly increased the mean number of inlier matches across most test sequences, with the greatest improvements generally obtained from the SumLog and SumLog-E transformations.
Sequences \texttt{InTheDark/0006} and \texttt{InTheDark/0071} are exceptions in that the standard Gray transformation performed best.
These sequences were recorded under similar conditions to the ``Map'' sequence, so feature matching can be expected to perform optimally on Gray images.
We saw little improvement in match counts on the \texttt{RobotCar/}Overcast-Night experiment, which we attribute to motion blur in the nighttime images making feature matching exceptionally difficult.
In contrast, the {MLP-E} method more than doubled the mean number of feature matches in the Sunny experiment.

We note that the pairwise encoder did not confer any significant benefit on the \texttt{VKITTI} sequences.
These results are consistent with \cite{Clement2017-gx,Corke2013-hl,McManus2014-op,Paton2017-fi}, where one or two sets of parameter values were sufficient to achieve good performance across varying daytime conditions.
In contrast, the encoder network provided a noticeable performance boost on most \texttt{InTheDark} and \texttt{RobotCar} sequences.
We attribute this difference to more variation in illumination and terrain, as a single transformation is less likely to perform well under more varied conditions.
We also note that the MLP-E transformation frequently performed similarly to the SumLog-E transformation, suggesting that, in spite of key assumptions being broken, a linear combination of log-responses as proposed by~\cite{Ratnasingam2010-cr} may in fact be an optimal solution for this problem, and that a careful choice of weights is the key to obtaining good cross-appearance feature matching over day-night cycles.

\subsection{Impact on Localization Performance}
We evaluate localization performance in an autonomous route-following context by examining the maximum distances in each sequence that would have been navigated using dead reckoning (e.g., visual odometry) as a result of failing to localize against the map.
These results are summarized in \Cref{tab:inthedark:loc_succ} for thresholds of 10, 20, and 30 inlier feature matches against the ``Map'' sequence.
A typical criterion for requiring manual intervention is dead reckoning in excess of 10 meters, depending on the accuracy of the underlying dead reckoning system.
Based on a relatively conservative threshold of 20 inlier feature matches against the ``Map'' sequence, we see that \texttt{InTheDark/0027} (evening) and \texttt{InTheDark/0041} (night) presented significant difficulty for localization, which was substantially alleviated using any of the four image transformations.
In particular, the SumLog-E transformation yielded maximum dead reckoning distances below four meters across all illumination conditions.
We see similar improvements using the SumLog-E method with a threshold of 10 inliers on the \texttt{RobotCar} dataset.
Together, these results imply that near-continuous 6-dof visual localization over a full day-night cycle is achievable using only a single mapping experience and a simple image pre-processing step, representing a dramatic reduction in data requirements to scale experience-based localization to long deployments.

With more conservative thresholds (e.g., 30 inliers), localization failures became more common, as expected.
However, the proposed image transformations continued to provide substantially more robust localization performance compared to standard grayscale images.
For example, we achieved a maximum distance on dead reckoning of 10.0 m using the SumLog-E method on sequence \texttt{InTheDark/0041} with a threshold of 30 inliers.
In contrast, using the standard Gray method would have required the system to rely continuously on dead reckoning for approximately 40\% of the route.

\section{Conclusions and Future Work}
This paper presented a method for learning pixel-wise RGB-to-grayscale colorspace mappings that explicitly maximize the number of inlier feature matches for a given input image pair, feature detector/matcher and operating environment.
Our key insight is that by training a deep neural network to predict the performance of a conventional non-differentiable feature detector/matcher, we can define a fully differentiable loss function that can be used to learn image transformations optimized for localization performance.
We evaluated our approach using both physically motivated and data-driven transformations and demonstrated substantially improved feature matching and localization performance on synthetic and real long-term vision datasets exhibiting severe illumination change, allowing experience-based localization to scale to long deployments with dramatically fewer bridging experiences.
We consistently achieved the best performance using a physically motivated weighted sum of log-responses with weights derived from a pairwise context encoder network.
In future work we plan to explore alternative loss functions such as pose estimation error, the use of feature locations or photometric consistency as a more granular supervision signal, and the impact of context feature dimension on MLP-based transformations.
We also intend to investigate the impact of our method on feature matching across seasonal appearance change.

\bibliographystyle{ieeetr}
\bibliography{refs}

\end{document}